\documentclass[runningheads]{llncs}
\usepackage[T1]{fontenc}
\usepackage{graphicx}
\usepackage{multirow}
\usepackage{amsmath}
\usepackage{bbding}
\begin{document}
\title{Saliency Guided Optimization of Diffusion Latents}


\author{Xiwen Wang\inst{1} \and
Jizhe Zhou\inst{1} \and
Xuekang Zhu\inst{1} \and
Cheng Li\inst{1} \and
Mao Li\inst{1}\inst{(}\Envelope\inst{)}
}

\authorrunning{F. Author et al.}
%
\institute{College of Computer Science, Sichuan University,
\\Chengdu, People’s Republic of China \\
\email{limao@scu.edu.cn}
}

\maketitle              
\begin{abstract}
With the rapid advances in diffusion models, generating decent images from text prompts is no longer challenging. The key to text-to-image generation is how to optimize the results of a text-to-image generation model so that they can be better aligned with human intentions or prompts.
Existing optimization methods commonly treat the entire image uniformly and conduct global optimization. These methods overlook the fact that when viewing an image, the human visual system naturally prioritizes attention toward salient areas, often neglecting less or non-salient regions. That is, humans are likely to neglect optimizations in non-salient areas. Consequently, although model retaining is conducted under the guidance of additional large and multimodality models, existing methods, which perform uniform optimizations, yield sub-optimal results. 
To address this alignment challenge effectively and efficiently, we propose Saliency Guided Optimization Of Diffusion Latents (\textbf{SGOOL}). We first employ a saliency detector to mimic the human visual attention system and mark out the salient regions. To avoid retraining an additional model, our method directly optimizes the diffusion latents. Besides, SGOOL utilizes an invertible diffusion process and endows it with the merits of constant memory implementation. Hence, our method becomes a parameter-efficient and plug-and-play fine-tuning method. Extensive experiments have been done with several metrics and human evaluation. Experimental results demonstrate the superiority of SGOOL in image quality and prompt alignment. 


\end{abstract}

%
\section{Introduction}
Text-to-image generation has been an unceasing task, with profound milestones achieved by Denoising Diffusion Models (DDMS) \cite{ho2020denoising}, such as Stable Diffusion \cite{rombach2022high}, DALL-E2 \cite{ramesh2022hierarchical} and Imagen \cite{saharia2022photorealistic}. To improve the quality of generation or meet the requirements of different downstream tasks, numerous methods \cite{ho2022classifier,hu2021lora,ruiz2023dreambooth,zhang2023adding} involve fine-tuning some layers of the diffusion model or training an auxiliary model using a different dataset. The critical idea of these methods lies in retraining under the guidance of additional large and multimodality models. The fine-tuning methods based on various guidance often require retraining an additional model, such as classifiers, to transform it into a noise-aware model. The gradients of this model are leveraged to guide the generation process of a diffusion model. Consequently, these methods result in their models are high in time expense and require significant computational resources.

\begin{figure}[!htb]\label{title_img}
\centering
\includegraphics[width=0.8\textwidth]{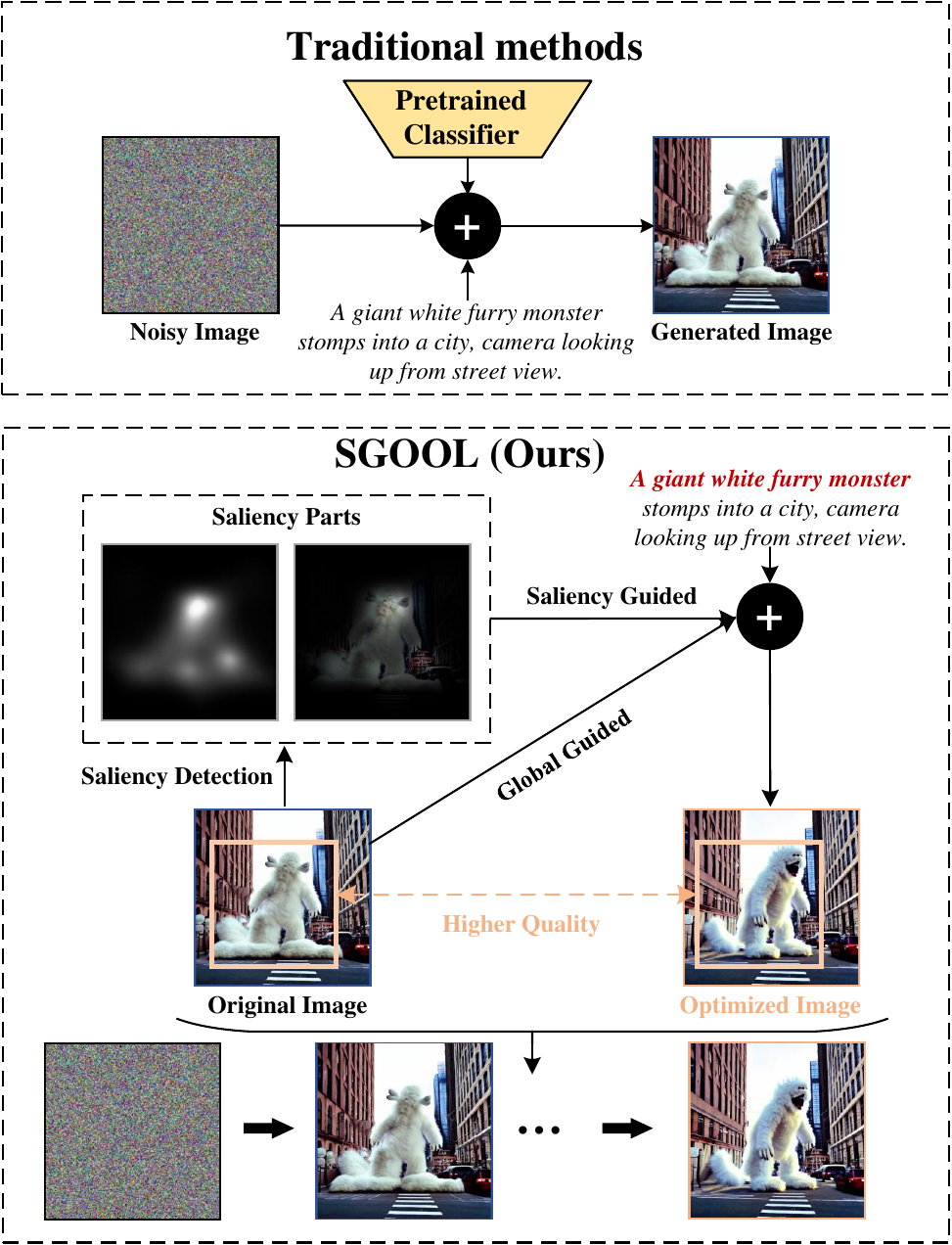}
\caption{Traditional methods vs. SGOOL. Given a prompt: "A giant white furry monster
stomps into a city, camera looking up from street view.". SGOOL generates an image with higher quality and more details.}
\end{figure}

Moreover, existing methods often consider the image as a whole without distinguishing different parts of the image. These methods overlook the fact that when humans appreciate an image, the visual system naturally prioritizes attention toward those objects or regions that most attract them, i.e., salient areas, often neglecting less or non-salient regions. Consequently, humans tend to grasp the theme, content, and emotion of paintings from saliency parts. This human trait is inherent in the human visual system \cite{bae2016novel,moorthy2011visual,sheikh2006image}. However, existing text-to-image methods ignore this human visual habit and typically optimize the whole image from a global perspective. As a result, these methods perform uniform optimizations, yielding sub-optimal results, and the quality of the generated images needs to be sufficiently improved.


Motivated by the above facts, we propose a novel diffusion model fine-tuning method in text-to-image tasks called SGOOL that stresses the salient regions during optimization. The salient regions are essential in conveying the meaning of the prompt and improving the overall image quality. As shown in Fig.~\ref{title_img}, the prompt suggests that the user wants the image to display a giant white furry monster. The traditional optimization method, taking Stable Diffusion with classifier guidance for example, considers the alignment between the global image and the prompt as guidance. As a result, the generated image does not effectively convey the prompt, nor does it satisfy the user. We introduce saliency guidance in addition to global guidance that is identical to traditional methods. First, we leverage a saliency detector to mimic the human visual system and mark out the salient regions. These regions are used as a mask for the original image to obtain specific saliency parts of the image. We consider the alignment between the saliency parts and the prompt as the saliency guidance to prioritize the saliency parts during optimization. Then, we combine the saliency guidance and global guidance to optimize the diffusion latents directly, which enables plug-and-play guidance without any retraining process. Besides, SGOOL achieves memory-efficient backpropagation by employing an invertible diffusion process.

Overall, the contributions of this study can be summarised as follows:
\begin{itemize}
    \item[$\bullet$] Novel perspectives in this study are brought to the field of image generation, especially in considering human visual preferences and saliency perception, opening up novel directions for future research.
    \item[$\bullet$] A saliency-aware and plug-and-play fine-tuning method is proposed to optimize the generated image without any retraining process.
    \item[$\bullet$] Experimental results demonstrate the effectiveness of SGOOL in both image quality and prompt alignment.
\end{itemize}

\section{Related work}
\subsection{Image Diffusion}
The Image Diffusion Model, initially proposed by Sohl-Dickstein et al. \cite{sohl2015deep}, has been utilized for image generation. Ho et al. \cite{ho2020denoising} established a link between the diffusion model and the denoising score matching model \cite{song2019generative} and introduced Denoising Diffusion Probabilistic Models (DDPM) as an alternative way to achieve superior image generation quality. Rombach et al. proposed the Latent Diffusion Models (LDM) \cite{rombach2022high}, which execute diffusion steps in the latent space, thereby decreasing computational costs. Dhariwal et al. \cite{dhariwal2021diffusion} enhanced image generation quality over GANs by identifying a superior model architecture and introducing a novel sampling technique, which is classifier guidance. However, classifier guidance necessitates training an additional classifier network. Ho et al. \cite{ho2022classifier} proposed a framework that trains a conditional model and an unconditional model jointly to provide classifier-free guidance with the same effect as classifier guidance. Nichol et al. \cite{nichol2021glide} proposed GLIDE, which considered CLIP \cite{radford2021learning} guidance as a guiding strategy and discussed its possibility of improving the generated image quality. Ramesh et al. \cite{ramesh2022hierarchical} proposed DALL-E2 to generate images from CLIP latent embeddings. Saharia et al. \cite{saharia2022photorealistic} proposed Imagen, which improves both sample fidelity and text-image alignment by increasing the size of the text encoder. In this study, we utilize Stable Diffusion model as our foundational framework to explore diffusion latents optimization guided by saliency to enhance the quality of the generated images. 

\subsection{Fine-tuning Methods} 
Fine-tuning is a technique in machine learning and deep learning that is often leveraged to improve the performance of a model on a specific task. Low-Rank Adaptation (LoRA) \cite{hu2021lora} is leveraged to efficiently fine-tune the diffusion model to adapt to specific tasks or domains. It reduces the number of trainable parameters required for downstream tasks by freezing the weights of the pre-trained model and injecting trainable low-rank decomposition matrices into layers of the model architecture. Nichol et al. \cite{nichol2021improved} discussed the scaling of the initial weight of the convolution layer in the diffusion model and proposed a "zero-module" to improve the training effect. HyperNetwork \cite{ha2016hypernetworks} originates from the Natural Language Processing (NLP) community and is an effective method to train a small recurrent neural network to enhance the origin network. HyperNetwork has also been applied to the diffusion model \cite{novelaiNovelAIImprovements}. DiffFit \cite{xie2023difffit} only fine-tunes bias terms and newly added scaling factors at a specific layer, but greatly speeds up training and reduces model storage costs. SVDiff \cite{han2023svdiff} reduces the risk of overfitting and linguistic drift by fine-tuning the singular values of the weight matrix to obtain a compact and efficient parameter space. While all of these fine-tuning methods achieve desirable results, they often neglect the salient regions in images. On the contrary, our method considers the optimisation of both the global image and the saliency region and can generate higher quality images with more details.

\subsection{Saliency Detection}
Saliency detection aims at identifying and segmenting salient objects in natural scenes that most attract human visual attention. Early saliency detection models primarily depended on the extraction of feature maps such as intensity, color, and orientation in a scale space, which were subsequently combined after performing normalization \cite{hou2007saliency,zhang2008sun}. The prominence of Convolutional Neural Networks (CNNs) \cite{lecun1998gradient} has helped to improve the performance of saliency models \cite{han2018salnet,jia2020eml,lou2022transalnet}. 
\section{Proposed Method}
\begin{figure}[!htb]
\centering
\includegraphics[width=0.9\textwidth]{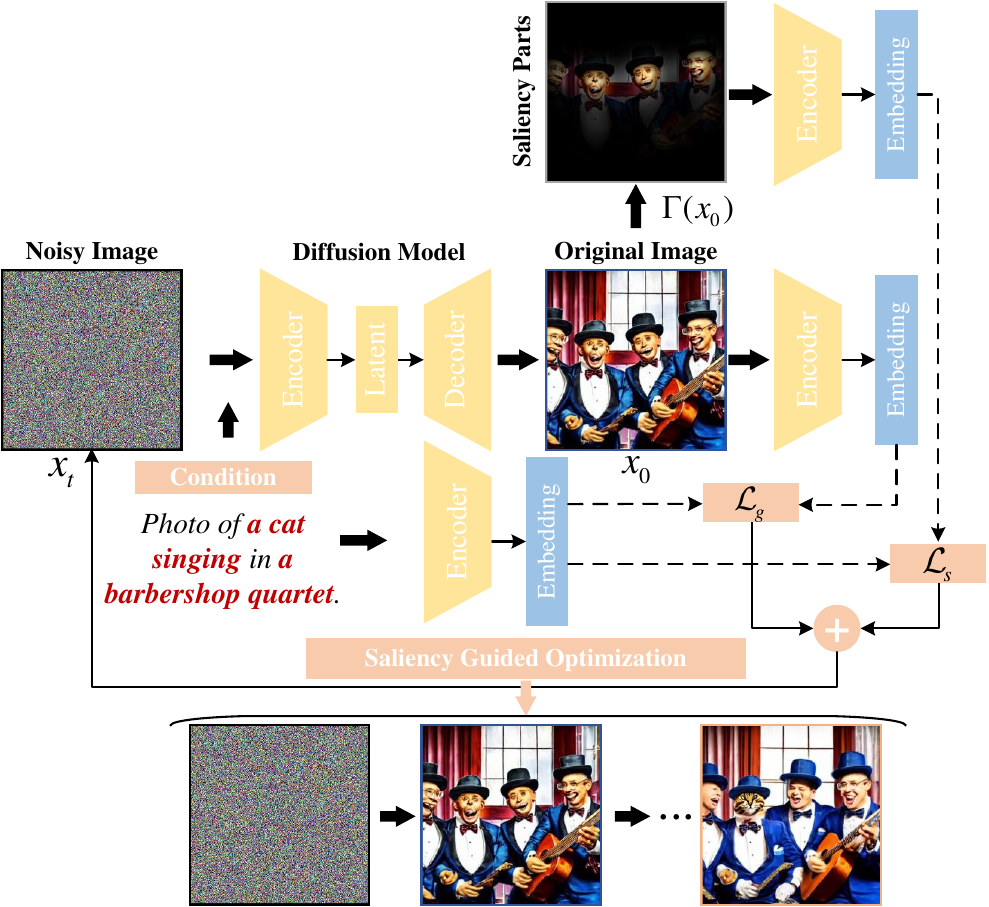}
\caption{The overall pipeline of SGOOL. SGOOL includes image generation, saliency detection, and optimizing steps. SGOOL obtains the saliency parts through the saliency detection model $\Gamma$. Then, both the global image and saliency parts are considered for optimization to improve the quality of the image.} \label{figure_pipeline}
\end{figure}
The architecture of our method is shown in Fig.\ref{figure_pipeline}. SGOOL mainly comprises image generation, saliency detection, and optimizing steps, but our method is far from employing the saliency detector to the diffusion model.

To avoid retraining an existing pretrained model, we aim to optimize the diffusion latents directly. Using the Denoising Diffusion Implicit Models (DDIM) \cite{song2020denoising} sampling method, the generated image $x_0$ is obtained by applying a series of denoising functions, which are conditioned on auxiliary information $c$ and varying timesteps $t$:
\begin{equation} \label{t20}
    x_0 = x_T [\Phi_{(c,T)} \circ \Phi_{(c,T-1)} \circ \dots \circ \Phi_{(c,0)}],
\end{equation}
where $\Phi$ denotes the denoising network of the diffusion model. Hence, it is mathematically rational that we optimize $x_T$ directly to obtain the better $x_0$. 

However, this process involves multiple applications of the denoising network $\Phi$, which means all the intermediate states need to be cached. 
In order to save memory, previous studies considered leveraging Invertible Neural Networks (INNs) to reduce memory costs. INNs enable the model to reconstruct the intermediate states during the reverse process by employing invertible $\Phi_{c,t}$ in Eq.~\eqref{t20}.
The recently proposed Direct Optimization Of Diffusion Latents (DOODL) \cite{wallace2023end} utilized an invertible diffusion process \cite{wallace2023edict} and performed end-to-end optimization of diffusion latent. Thus, we apply DOODL to our model to optimize the latent variables directly. However, DOODL treats the entire image uniformly and conducts global optimization, leading to sub-optimal results.

Due to the limited attention of human eyes, people tend to allocate more attention to certain areas of the image while neglecting others. This inspiration leads us to consider assigning different weights to different parts of the image in the image generation task. We employ TransalNet \cite{lou2022transalnet} as our saliency detection model $\Gamma$ to extract the salient regions from the image. This involves feeding the image into the model, which then identifies the areas of interest, denoted as $\Gamma(x_0)$. Subsequently, we extract and crop these salient regions from the original image to generate the final image saliency parts $S$ that are perceptible to humans. 

To facilitate the computation of the differences between text and image, we utilize the CLIP model to unify the input prompt and image. The loss function is designed by computing the spherical distance loss between the input prompt embedding and the image embedding. This loss can be interpreted as the similarity between the text and the image \cite{wallace2023end}. The spherical distance loss is computed by
\begin{equation} \label{distance}
    d(x,y) = 2\lambda\sqrt{sin^{-1}(\frac{\Vert{x-y}\Vert}{2})}.
\end{equation}

Based on the previous discussion, we generate high-quality images in this study by co-optimizing the saliency parts $S$ and the overall image $G$. In detail, the distance between the embedding of the generated image and the input prompt is regarded as the optimization objective. The embedding of the input prompt $C$ by the CLIP text encoder, the embedding of the global image $G$ by the CLIP image encoder, and the embedding of the saliency parts $S$ of $G$ by the CLIP image encoder are denoted as $Emb_c$, $Emb_s$, and $Emb_g$ respectively. Specifically, we design the following two losses.
\begin{equation} \label{salient loss}
    \mathcal{L}_s = 2\lambda\sqrt{sin^{-1}(\frac{\Vert{Emb_c-Emb_s}\Vert}{2})},
\end{equation}
\begin{equation} \label{holistic loss}
    \mathcal{L}_g = 2\lambda\sqrt{sin^{-1}(\frac{\Vert{Emb_c-Emb_g}\Vert}{2})}.
\end{equation}

The above two losses correspond to the detail quality and overall quality of the image. Thus, their proportion in the total loss function should be carefully balanced. Driven by this, we deploy a loss design strategy that allows the two types of loss to be adaptable. The final loss function is designed by
\begin{equation} \label{total loss}
    \mathcal{L} = \alpha\mathcal{L}_s + (1-\alpha)\mathcal{L}_g.
\end{equation}

In this way, the final loss function $\mathcal{L}$ regards the relationships between saliency parts and the global image simultaneously, which helps to balance local details and global consistency in the generation process. This saliency-aware loss is leveraged to optimize image latent. The gradients are computed on the noised latent $x_t$ and leveraged to enhance the conditioning effect of the input prompt on both salient and global aspects of the original generated image.  

\section{Experiments}

In this section, we compare SGOOL with vanilla Stable Diffusion (SD) and Stable Diffusion with CLIP guidance (Baseline). Then, we present a detailed analysis of the experimental results.

\subsection{Experiment setup}

\subsubsection{Datasets}
We evaluate SGOOL on three public datasets, including CommonSyntacticProcesses (CSP) \cite{lee2024holistic}, DailyDallE \cite{lee2024holistic}, and DrawBench \cite{lee2024holistic}. CSP contains 52 prompts from 8 different grammatical phenomena to evaluate the comprehension of different grammatical prompts and the image generation capability of the model. The DailyDallE contains 99 prompts, most of which are complex. These prompts come from a featured artist on OpenAI's blog post to evaluate the image generation capability of the model under complex prompts. The DrawBench contains 200 prompts from 11 categories, a comprehensive and challenging set of prompts that support the evaluation and comparison of text-to-image models.

\subsubsection{Implementation details}
In SD, Baseline, and SGOOL, we leverage the CLIP model based on ViT/B-32 to generate the image embedding and the text embedding. All models are based on the same diffusion model, Stable Diffusion V1.4. During the experiments, the same prompt is provided and the same random seed is set to ensure the fairness of the experiments. The output image sizes of the models are all $256 \times 256$. For the saliency detection model TransalNet leveraged in SGOOL, the default weights and configurations provided by it are used. We use $\alpha = 0.5$ in $\mathcal{L}$.

\subsubsection{Evaluation metrics}
How to evaluate the text-to-image generation model has been a tricky challenge \cite{zhang2023text}. In this study, we take the image quality and text-image alignment as the evaluation metric, which reflects the fidelity of the images and whether the generated images match the text semantic content well. CLIP score is widely applied to measure the alignment of text-image images. CLIP score evaluates the alignment between the input prompt and the generated image by computing the cosine similarity between the prompt embedding and the image embedding generated by CLIP \cite{hessel2021clipscore}. We compute the CLIP scores of different models based on ViT-B/32. We evaluate human preference for the generated images by Human Preference Score (HPS) \cite{wu2023human}. HPS provides an objective evaluation metric based on human preferences.
In this study, we use the default weights and configurations of HPS to evaluate the experimental results.

We also evaluate the natural human preference for the generated images through subjective human perception tests. We recruit 100 volunteers to perform image quality and semantic consistency tests. The image quality test is designed to evaluate the detail and the overall aesthetic quality of the generated image. The semantic consistency test evaluates the semantic consistency between the generated image and the prompt. In the test, participants are asked to evaluate 100 groups of images. Each group contains three randomly selected images from the generated images of SD, Baseline, and SGOOL, respectively. Participants are given unlimited time to select the image they think is the most realistic and consistent with the prompt. 


\subsection{Performance Evaluation}

In this section, we present the quantitative and qualitative results of SD, Baseline, and SGOOL separately. First, we compare the CLIP score and HPS results of three models on different datasets. Subsequently, the human subjective visual evaluations of the 3 models are compared, which include image quality and semantic consistency.

\subsubsection{Quantitative Results}  

Table~\ref{tab_quantitative} presents the CLIP score and HPS results for SD, Baseline, and our model on all three datasets: CSP, DailyDallE, and DrawBench.
The CLIP score and HPS values are in the range of $[0,100]$ and $[0,1]$, respectively. In both cases, a higher score means a better model. As can be seen from the table, our model significantly outperforms SD and Baseline on all datasets under both CLIP score and HPS metrics. The average results of our model on CLIP score and HPS are $3.05$ and $0.0029$ higher than the second place, respectively.

\begin{table*}[!htb] 
\centering
\caption{Quantitative results for CLIP score and HPS on CSP, DailyDallE, and DrawBench datasets. }
\label{tab_quantitative}
\resizebox{\textwidth}{!}{
\begin{tabular}{c|ccc|c|ccc|c}
\hline
\multicolumn{1}{c|}{\multirow{2}{*}{Metrics}} & \multicolumn{4}{c|}{{CLIP Score}}                                   & \multicolumn{4}{c}{{HPS}}                                               \\ \cline{2-9} 
\multicolumn{1}{c|}{}                         & CSP            & DailyDallE     & DrawBench      & {Average}        & CSP             & DailyDallE      & DrawBench       & {Average}         \\ \hline
SD                                            & 31.67          & 31.67          & 32.15          & 31.83          & 0.2650          & 0.2573          & 0.2595          & 0.2606          \\
Baseline                                      & 31.61          & 35.20          & 31.92          & 32.91          & 0.2705          & 0.2622          & 0.2627          & 0.2651          \\
\textbf{SGOOL}                                 & \textbf{34.84} & \textbf{38.11} & \textbf{34.93} & \textbf{35.96} & \textbf{0.2778} & \textbf{0.2630} & \textbf{0.2632} & \textbf{0.2680} \\ \hline
\end{tabular}}
\end{table*}

\begin{figure}[!htb]
\centering
\includegraphics[width=1\textwidth, trim=3cm 1cm 3cm 2cm,clip]{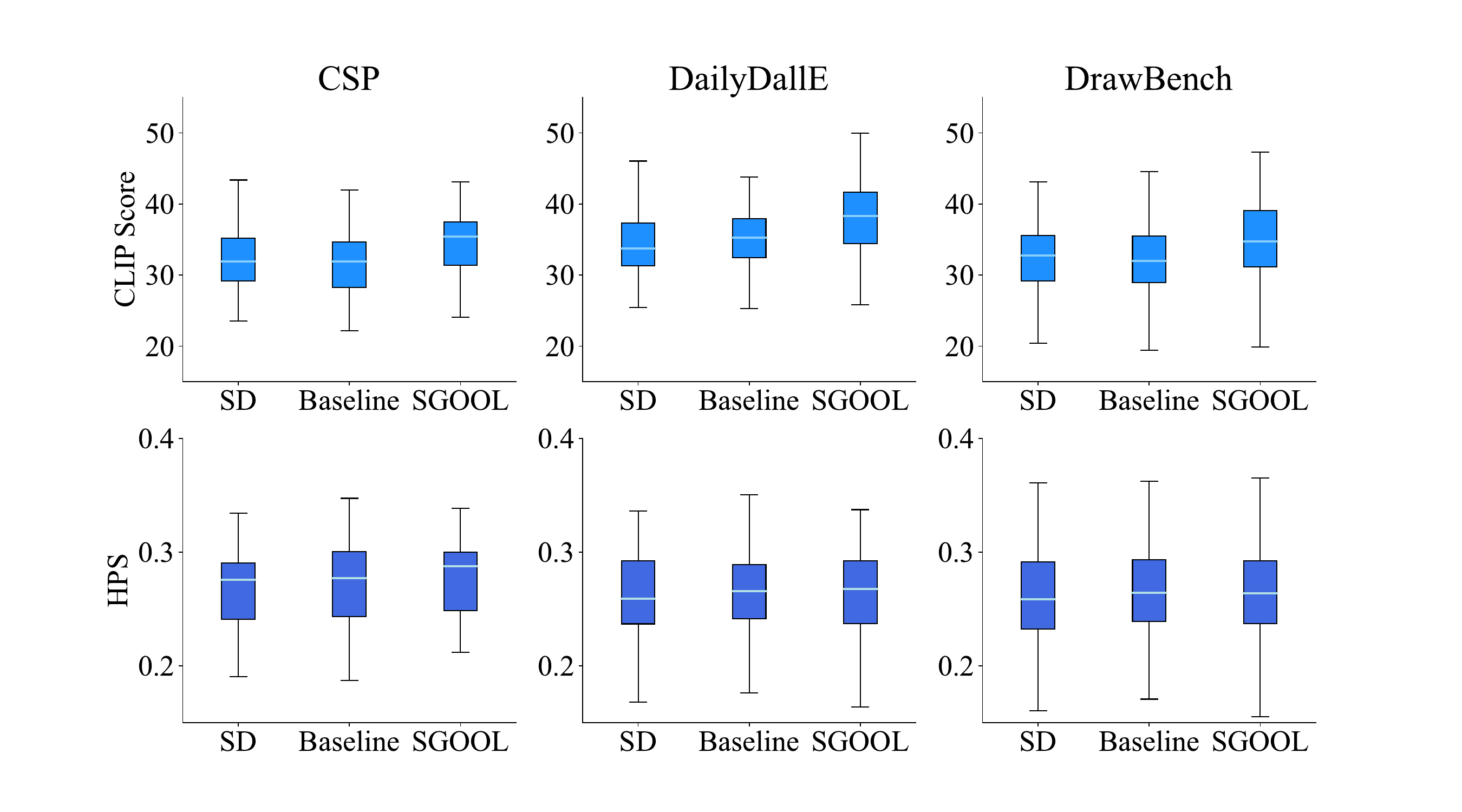}
\caption{Box plots for CLIP score and HPS.} \label{fig_boxplots}
\end{figure}

We plot the box plots of CLIP score and HPS with respect to SD, Baseline, and SGOOL on different datasets, as shown in Fig.~\ref{fig_boxplots}. It can be seen that our model outperforms the other models, indicating that our model is more capable of generating images that are consistent with the prompts. However, in the box plot, it is not easy to visualize the comparison from the box plot due to the size of this evaluation metric at $[0,1]$. Therefore, we proceed to plot the corresponding bar plots. It can be seen that SGOOL outperforms SD and Baseline on all datasets under both CLIP score and HPS metrics. The quantitative results demonstrate that our model can generate more semantically consistent and human-preferred images. 

The vanilla Stable Diffusion model employs no fine-tuning techniques, resulting in poor overall performance. In contrast, the Baseline model introduces CLIP guidance to generate more vivid images. Although the Baseline model does improve the image quality, it ignores the salient regions in the image. Unlike the previous two, SGOOL emphasizes optimizing the latents by focusing on the salient regions of the image to improve the image quality. Besides, to avoid over-focusing on salient regions leading to discontinuities in the local image structure, SGOOL also considers global image information to improve overall image quality.

\begin{figure}[!htb]
\centering
\includegraphics[width=1\textwidth, trim=3cm 0cm 3cm 1cm,clip]{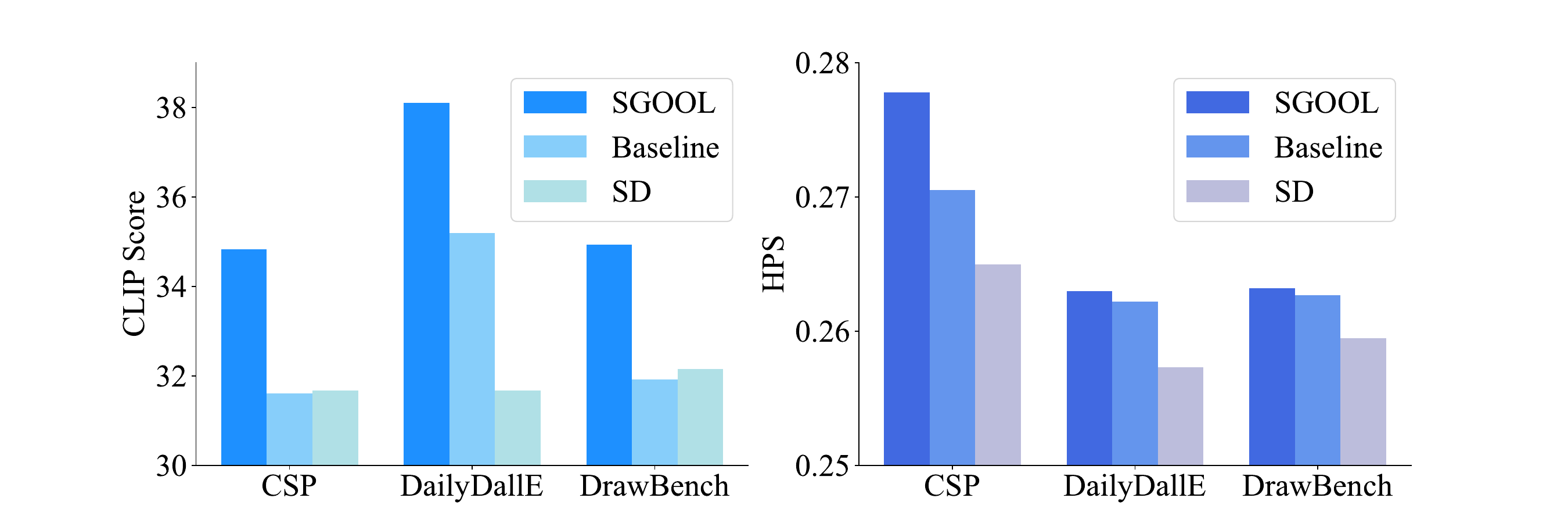}
\caption{Bar plots for CLIP score and HPS.} \label{fig_barplots}
\end{figure}

\subsubsection{Qualitative Results}  
\begin{figure}[!htb]
    \centering
    \includegraphics[width=1\textwidth]{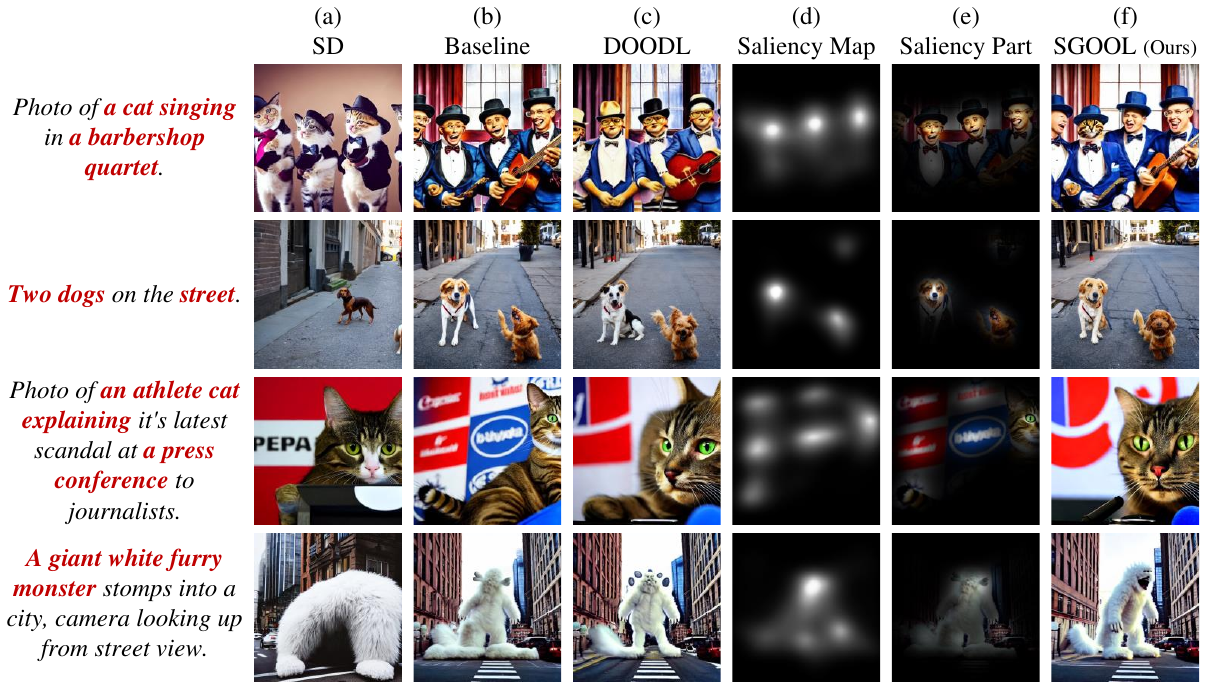}
    \caption{Examples of images generated by (a) vanilla Stable Diffusion (SD), (b) Stable Diffusion with CLIP guidance (Baseline), (c) DOODL, and (f) SGOOL (Ours) from the same random seed.}
    \label{fig_exp_1}
\end{figure}

\begin{figure}[!htb]
    \centering
    \includegraphics[width=1\textwidth]{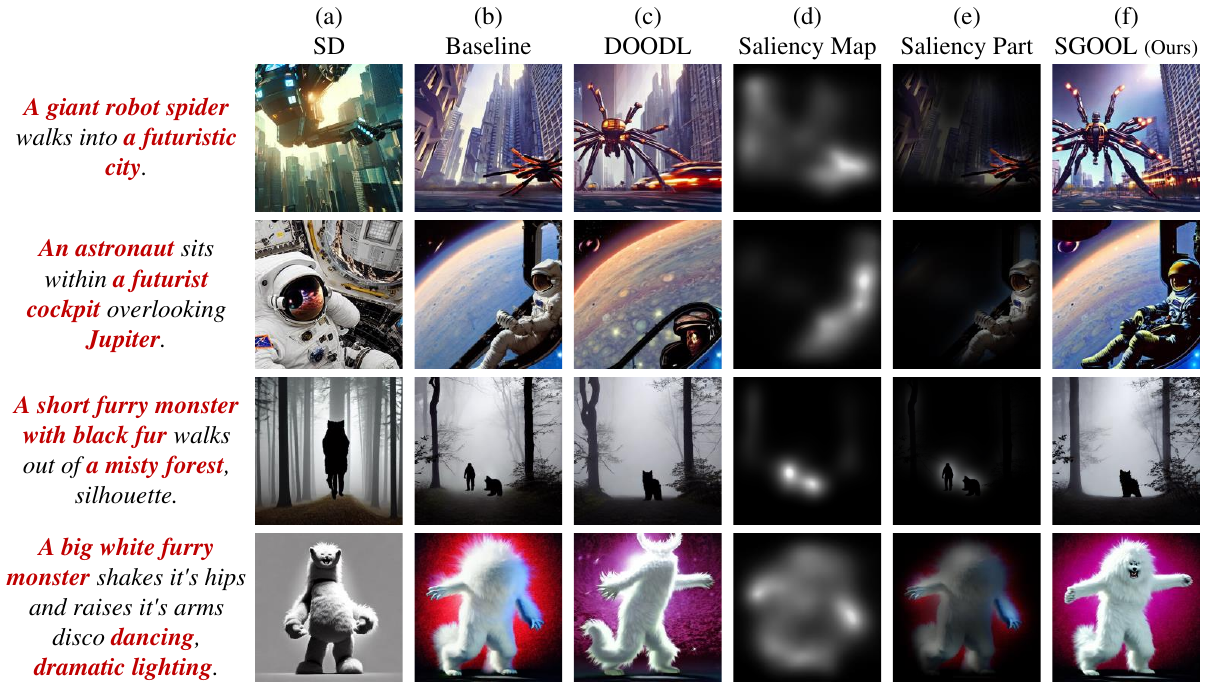}
    \caption{Examples of images generated by (a) vanilla Stable Diffusion (SD), (b) Stable Diffusion with CLIP guidance (Baseline), (c) DOODL, and (f) SGOOL (Ours) from the same random seed.}
    \label{fig_exp_2}
\end{figure}

We perform a qualitative comparison by visually comparing the quality of images generated by SGOOL, DOODL, Baseline, and SD with the same prompts and seeds. The number of optimizations is set to 50 for both SGOOL and DOODL. Fig.~\ref{fig_exp_1} and Fig.~\ref{fig_exp_2} present the input prompts, saliency maps, saliency parts, and corresponding images generated by SGOOL, DOODL, Baseline, and SD with the same prompts and seeds.

In the first row of the prompts in Fig.~\ref{fig_exp_1}, the subjects of the prompt are "a cat singing" and "a barbershop quartet". There are four cats in the image generated by SD, and the content of the image is poorly aligned with the prompt. The cat is ignored in the image generated by Baseline, and there is a lack of detail in the portrayal of the face and the details in the image. DOODL attempts to generate an image that is consistent with the prompt. However, since DOODL optimizes the global image directly, the persons in the image are optimized toward the cat. Unlike DOODL, SGOOL focuses on optimizing the salient areas of the image, generating images that are more consistent with the prompt and of higher quality. As can be seen from the corresponding saliency images, the saliency parts of the image lie in the head areas. SGOOL focuses on optimizing the part of the image that the human eye focuses on first and takes into account the coordination between the local and global images. This results in a final generated image that is more favorable to the human eyes and consistent with the input prompt. The generated images in the subsequent rows similarly demonstrate the superiority of SGOOL in both image quality and prompt alignment.

In summary, while DOODL optimizes directly on diffuse latents, its optimization considers the global image and ignores the importance of the saliency parts for optimization. In contrast, SGOOL considers both the global image and saliency parts of the generated image. SGOOL focuses on overall consistency and pays special attention to optimizing details and local features. This includes accurately depicting details such as morphology, expression, and texture. Images generated by SGOOL significantly improve image quality and semantic consistency with prompts. The saliency images show the critical regions of the image, which are usually where visual attention is focused. By generating images and saliency images, we can observe that SGOOL pays special attention to the salient regions in the image while generating the image and hence can optimize the fine-grained features in the image, which helps it to generate higher quality images.

\begin{figure}[!htb]
\centering
\includegraphics[width=0.6\textwidth]{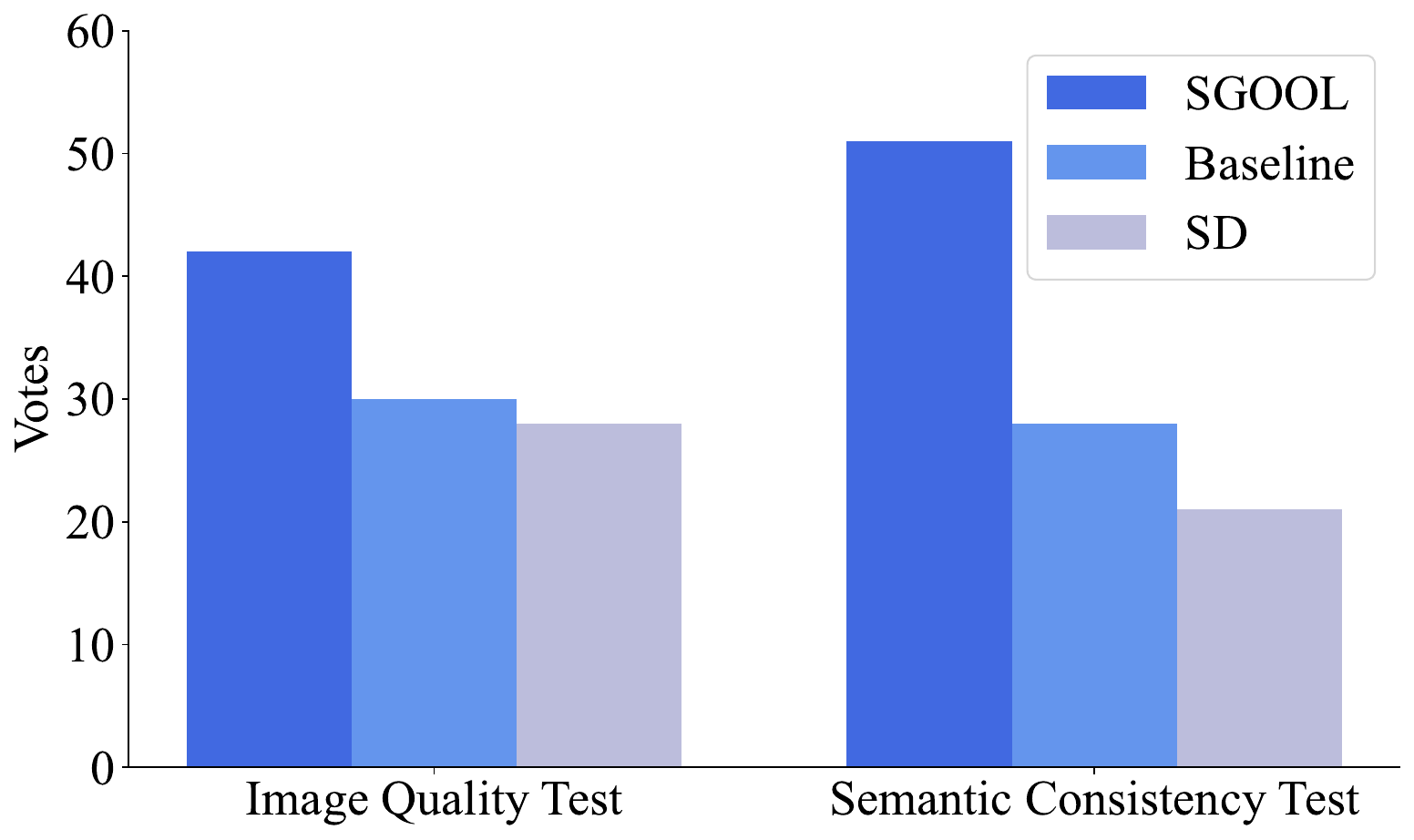}
\caption{Bar plots for human perception test results. A higher value means more preferred by volunteers for image quality (referring to both the detail of the image and the overall aesthetic quality of the image) or semantic consistency.} \label{fig_human_test}
\end{figure}



The human perception test is next, which follows the previously described. We test images generated on all prompts, and after the experiment is completed, we count the votes of participants on all images generated by the three models. The final results of the experiments are shown in Fig.~\ref{fig_human_test}. As can be seen from the figure, SGOOL is preferred for both image quality and semantic consistency over Baseline and SD.

\section{Conclusion}
In this study, we start from the perspective of artistic creation to make end-to-end modifications to the diffusion model. Motivated by this, we utilize the saliency detector to enable the diffusion model to "see" areas humans pay more attention to and optimize with global guidance and saliency guidance. The experimental results demonstrate that SGOOL improves the quality of the generated images with great detail.  Furthermore, we hope that our unique perspective in this study can inspire a broader audience to consider image generation from the perspective of artistic creation.

\begin{credits}
\subsubsection{\ackname} This work is supported by the Digital Media Art, Key Laboratory of Sichuan Province, Sichuan Conservatory of Music (Grant No. 22DMAKL04); the  Natural Science Foundation of Sichuan (No. 2023NSFSC0474).

\subsubsection{\discintname}
\textbf{The authors have no competing interests to declare that are relevant to the content of this article.}

\end{credits}

%
%
%
\bibliographystyle{splncs04}
\bibliography{refer}

\begin{thebibliography}{10}
\providecommand{\url}[1]{\texttt{#1}}
\providecommand{\urlprefix}{URL }
\providecommand{\doi}[1]{https://doi.org/#1}

\bibitem{novelaiNovelAIImprovements}
Anlatan: {N}ovel{A}{I} {I}mprovements on {S}table {D}iffusion --- blog.novelai.net. \url{https://blog.novelai.net/novelai-improvements-on-stable-diffusion-e10d38db82ac}, [Accessed 16-05-2024]

\bibitem{bae2016novel}
Bae, S.H., Kim, M.: A novel image quality assessment with globally and locally consilient visual quality perception. IEEE Transactions on Image Processing  \textbf{25}(5),  2392--2406 (2016)

\bibitem{dhariwal2021diffusion}
Dhariwal, P., Nichol, A.: Diffusion models beat gans on image synthesis. Advances in neural information processing systems  \textbf{34},  8780--8794 (2021)

\bibitem{ha2016hypernetworks}
Ha, D., Dai, A., Le, Q.V.: Hypernetworks. arXiv preprint arXiv:1609.09106  (2016)

\bibitem{han2018salnet}
Han, L., Li, X., Dong, Y.: Salnet: Edge constraint based end-to-end model for salient object detection. In: Pattern Recognition and Computer Vision: First Chinese Conference, PRCV 2018, Guangzhou, China, November 23-26, 2018, Proceedings, Part IV 1. pp. 186--198. Springer (2018)

\bibitem{han2023svdiff}
Han, L., Li, Y., Zhang, H., Milanfar, P., Metaxas, D., Yang, F.: Svdiff: Compact parameter space for diffusion fine-tuning. In: Proceedings of the IEEE/CVF International Conference on Computer Vision. pp. 7323--7334 (2023)

\bibitem{hessel2021clipscore}
Hessel, J., Holtzman, A., Forbes, M., Bras, R.L., Choi, Y.: Clipscore: A reference-free evaluation metric for image captioning. arXiv preprint arXiv:2104.08718  (2021)

\bibitem{ho2020denoising}
Ho, J., Jain, A., Abbeel, P.: Denoising diffusion probabilistic models. Advances in neural information processing systems  \textbf{33},  6840--6851 (2020)

\bibitem{ho2022classifier}
Ho, J., Salimans, T.: Classifier-free diffusion guidance. arXiv preprint arXiv:2207.12598  (2022)

\bibitem{hou2007saliency}
Hou, X., Zhang, L.: Saliency detection: A spectral residual approach. In: 2007 IEEE Conference on computer vision and pattern recognition. pp.~1--8. Ieee (2007)

\bibitem{hu2021lora}
Hu, E.J., Shen, Y., Wallis, P., Allen-Zhu, Z., Li, Y., Wang, S., Wang, L., Chen, W.: Lora: Low-rank adaptation of large language models. arXiv preprint arXiv:2106.09685  (2021)

\bibitem{jia2020eml}
Jia, S., Bruce, N.D.: Eml-net: An expandable multi-layer network for saliency prediction. Image and vision computing  \textbf{95},  103887 (2020)

\bibitem{lecun1998gradient}
LeCun, Y., Bottou, L., Bengio, Y., Haffner, P.: Gradient-based learning applied to document recognition. Proceedings of the IEEE  \textbf{86}(11),  2278--2324 (1998)

\bibitem{lee2024holistic}
Lee, T., Yasunaga, M., Meng, C., Mai, Y., Park, J.S., Gupta, A., Zhang, Y., Narayanan, D., Teufel, H., Bellagente, M., et~al.: Holistic evaluation of text-to-image models. Advances in Neural Information Processing Systems  \textbf{36} (2024)

\bibitem{lou2022transalnet}
Lou, J., Lin, H., Marshall, D., Saupe, D., Liu, H.: Transalnet: Towards perceptually relevant visual saliency prediction. Neurocomputing  \textbf{494},  455--467 (2022)

\bibitem{moorthy2011visual}
Moorthy, A.K., Wang, Z., Bovik, A.C.: Visual perception and quality assessment. Optical and Digital Image Processing: Fundamentals and Applications pp. 419--439 (2011)

\bibitem{nichol2021glide}
Nichol, A., Dhariwal, P., Ramesh, A., Shyam, P., Mishkin, P., McGrew, B., Sutskever, I., Chen, M.: Glide: Towards photorealistic image generation and editing with text-guided diffusion models. arXiv preprint arXiv:2112.10741  (2021)

\bibitem{nichol2021improved}
Nichol, A.Q., Dhariwal, P.: Improved denoising diffusion probabilistic models. In: International conference on machine learning. pp. 8162--8171. PMLR (2021)

\bibitem{radford2021learning}
Radford, A., Kim, J.W., Hallacy, C., Ramesh, A., Goh, G., Agarwal, S., Sastry, G., Askell, A., Mishkin, P., Clark, J., et~al.: Learning transferable visual models from natural language supervision. In: International conference on machine learning. pp. 8748--8763. PMLR (2021)

\bibitem{ramesh2022hierarchical}
Ramesh, A., Dhariwal, P., Nichol, A., Chu, C., Chen, M.: Hierarchical text-conditional image generation with clip latents. arXiv preprint arXiv:2204.06125  \textbf{1}(2), ~3 (2022)

\bibitem{rombach2022high}
Rombach, R., Blattmann, A., Lorenz, D., Esser, P., Ommer, B.: High-resolution image synthesis with latent diffusion models. In: Proceedings of the IEEE/CVF conference on computer vision and pattern recognition. pp. 10684--10695 (2022)

\bibitem{ruiz2023dreambooth}
Ruiz, N., Li, Y., Jampani, V., Pritch, Y., Rubinstein, M., Aberman, K.: Dreambooth: Fine tuning text-to-image diffusion models for subject-driven generation. In: Proceedings of the IEEE/CVF Conference on Computer Vision and Pattern Recognition. pp. 22500--22510 (2023)

\bibitem{saharia2022photorealistic}
Saharia, C., Chan, W., Saxena, S., Li, L., Whang, J., Denton, E.L., Ghasemipour, K., Gontijo~Lopes, R., Karagol~Ayan, B., Salimans, T., et~al.: Photorealistic text-to-image diffusion models with deep language understanding. Advances in neural information processing systems  \textbf{35},  36479--36494 (2022)

\bibitem{sheikh2006image}
Sheikh, H.R., Bovik, A.C.: Image information and visual quality. IEEE Transactions on image processing  \textbf{15}(2),  430--444 (2006)

\bibitem{sohl2015deep}
Sohl-Dickstein, J., Weiss, E., Maheswaranathan, N., Ganguli, S.: Deep unsupervised learning using nonequilibrium thermodynamics. In: International conference on machine learning. pp. 2256--2265. PMLR (2015)

\bibitem{song2020denoising}
Song, J., Meng, C., Ermon, S.: Denoising diffusion implicit models. arXiv preprint arXiv:2010.02502  (2020)

\bibitem{song2019generative}
Song, Y., Ermon, S.: Generative modeling by estimating gradients of the data distribution. Advances in neural information processing systems  \textbf{32} (2019)

\bibitem{wallace2023end}
Wallace, B., Gokul, A., Ermon, S., Naik, N.: End-to-end diffusion latent optimization improves classifier guidance. In: Proceedings of the IEEE/CVF International Conference on Computer Vision. pp. 7280--7290 (2023)

\bibitem{wallace2023edict}
Wallace, B., Gokul, A., Naik, N.: Edict: Exact diffusion inversion via coupled transformations. In: Proceedings of the IEEE/CVF Conference on Computer Vision and Pattern Recognition. pp. 22532--22541 (2023)

\bibitem{wu2023human}
Wu, X., Hao, Y., Sun, K., Chen, Y., Zhu, F., Zhao, R., Li, H.: Human preference score v2: A solid benchmark for evaluating human preferences of text-to-image synthesis. arXiv preprint arXiv:2306.09341  (2023)

\bibitem{xie2023difffit}
Xie, E., Yao, L., Shi, H., Liu, Z., Zhou, D., Liu, Z., Li, J., Li, Z.: Difffit: Unlocking transferability of large diffusion models via simple parameter-efficient fine-tuning. In: Proceedings of the IEEE/CVF International Conference on Computer Vision. pp. 4230--4239 (2023)

\bibitem{zhang2023text}
Zhang, C., Zhang, C., Zhang, M., Kweon, I.S.: Text-to-image diffusion model in generative ai: A survey. arXiv preprint arXiv:2303.07909  (2023)

\bibitem{zhang2008sun}
Zhang, L., Tong, M.H., Marks, T.K., Shan, H., Cottrell, G.W.: Sun: A bayesian framework for saliency using natural statistics. Journal of vision  \textbf{8}(7),  32--32 (2008)

\bibitem{zhang2023adding}
Zhang, L., Rao, A., Agrawala, M.: Adding conditional control to text-to-image diffusion models. In: Proceedings of the IEEE/CVF International Conference on Computer Vision. pp. 3836--3847 (2023)

\end{thebibliography}
%




\end{document}